\documentclass[10pt,journal,compsoc,onecolumn]{IEEEtran}

\usepackage{amsmath}
\usepackage{amssymb}
\usepackage{mathtools}
\usepackage{amsthm}

\theoremstyle{plain}
\newtheorem{theorem}{Theorem}[section]

\newtheorem{lemma}[theorem]{Lemma}
\newtheorem{corollary}[theorem]{Corollary}
\theoremstyle{definition}
\newtheorem{definition}[theorem]{Definition}

\theoremstyle{remark}
\newtheorem{remark}[theorem]{Remark}

\usepackage[dvipsnames]{xcolor}
\usepackage{multirow}

\usepackage{colortbl}

\usepackage[textsize=small,colorinlistoftodos]{todonotes}

\usepackage{commath}

\usepackage{hyperref}
\hypersetup{
    colorlinks=true,
    linkcolor=blue,
    filecolor=magenta,      
    urlcolor=blue,
    pdftitle={Overleaf Example},
    pdfpagemode=FullScreen,
    citecolor=Blue
    }
\usepackage{amsfonts}
\usepackage{amsmath}
\usepackage{amssymb}
\usepackage{mathtools}
\usepackage{multicol}
\usepackage{multirow,tabularx}
\usepackage{booktabs}
\usepackage{caption}
\usepackage{enumitem}
\usepackage{subfigure}
\usepackage[dvipsnames]{xcolor}
\usepackage{hyperref}
\usepackage[para,online,flushleft]{threeparttable}

\ifCLASSOPTIONcompsoc
  \usepackage[nocompress]{cite}
\else
  \usepackage{cite}
\fi

\ifCLASSOPTIONcaptionsoff
    \usepackage[nomarkers]{endfloat}
    \let\MYoriglatexcaption\caption
    \renewcommand{\caption}[2][\relax]{\MYoriglatexcaption[#2]{#2}}
\fi

\usepackage{url}

\begin{document}

\title{Asynchronous Message Passing for Addressing Oversquashing in Graph Neural Networks}

\author{Kushal Bose and Swagatam Das\\
\IEEEcompsocitemizethanks{\IEEEcompsocthanksitem Electronics and Communication Sciences Unit, Indian Statistical Institute, Kolkata, India.\protect\\

E-mail: kushalbose92@gmail.com, swagatam.das@isical.ac.in
}
}

\IEEEtitleabstractindextext{%
\begin{abstract}
Graph Neural Networks (GNNs) suffer from oversquashing, where structural bottlenecks limit message propagation between distant nodes, hindering tasks that require long-range interactions. Existing remedies are limited: graph rewiring alters edge connectivity, compromising inductive bias, while increasing channel capacity adds parameters. In this work, we propose an efficient, model-agnostic framework that asynchronously updates node features across layers, unlike standard synchronous message passing. At each layer, only a centrality-guided batch of nodes updates, letting information propagate sequentially rather than compress simultaneously into fixed-capacity channels. We show theoretically that our framework's sensitivity bound decays more slowly with depth than synchronous message passing. We have applied the framework to six standard and two long-range graph classification benchmarks, and it achieves notable gains, including $5\%$ and $4\%$ improvements on REDDIT-BINARY and Peptides-struct, respectively.    
\end{abstract}

\begin{IEEEkeywords}Graph, Convolution, Message passing, Oversquashing, Centrality, Asynchronization, Oversmoothing
\end{IEEEkeywords}}

\maketitle

\section{Introduction}
Graph Neural Networks \cite{topo_gnn_survey}, \cite{survey_gnn_tnnls}, \cite{gnn}, \cite{sgc}, \cite{pna} predominantly aggregate localized information, which is insufficient for tasks requiring long-range interactions, like molecular property predictions \cite{benchmarking}, \cite{lrgb}. The interactions among multi-hop nodes \cite{mixhop} can be achieved by increasing the size of receptive fields. Yet, the increasing number of neighbors introduces the problem of compressing the growing amount of information in the fixed-dimensional vectors, which creates potential information bottlenecks. This phenomenon is commonly known as \textit{Oversquashing} \cite{oversquashing, oversquashing_wdt}, which causes significant information loss, particularly affecting signals from distant nodes, crucial for the downstream task \cite{drew}.  

\begin{figure*}[!ht]
    \centering
    \includegraphics[width=0.80\textwidth]{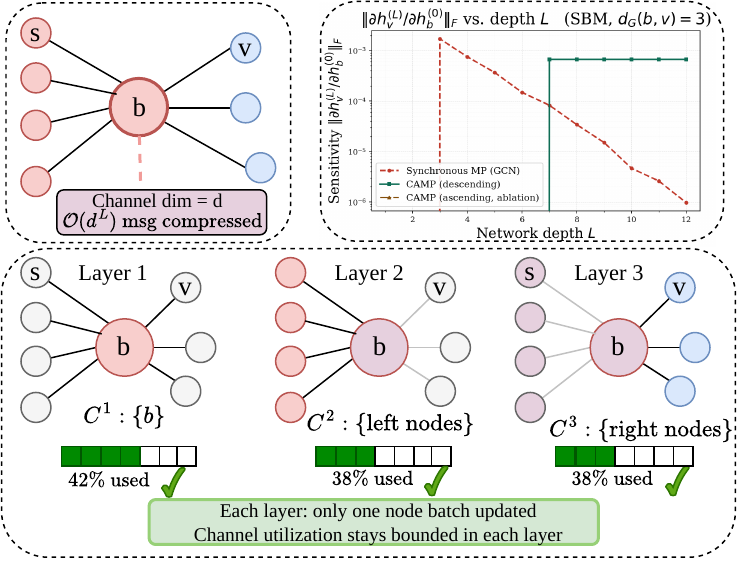}
    \caption{The comparative study of synchronous and asynchronous GNN is presented. In the synchronous model, all node features are updated simultaneously. The node feature capacity ultimately saturates while aggregating information from multi-hop neighbors. In the asynchronous framework, we first sort the node centrality values in descending order. For a $3$-layered architecture, the centrality set is divided into three disjoint subsets. In each layer, the features of nodes in the pertinent subset are updated. The node feature capacity never gets saturated while aggregating exponential information from growing neighborhoods. Evidently, the feature sensitivity remains steady for CAMP but degrades for synchronous models.  }
    \label{fig:camp_model}
\end{figure*}

\textbf{Limitations of Existing Approaches.} Mitigating oversquashing begins with converting the input graph into a fully connected (FA) graph \cite{oversquashing} and applying it to either the final layer or every layer of the base GNN model. Recent approaches have attempted to address oversquashing through graph rewiring methods \cite{sdrf}, \cite{fosr}, \cite{diffwire}, \cite{laser}, \cite{gtr}, \cite{prmpnn}, which add edges as "shortcuts" between long-distance nodes to facilitate multi-hop information flow. Nevertheless, FA or other rewiring methods tamper with the original graph topology, potentially compromising the inductive bias and incurring loss of information. As an alternative avenue, Graph Transformers (GTs) \cite{graphtransformer} have emerged, enabling global message passing between all pairs of nodes, but they introduce prohibitive computational complexity, particularly for large-scale graphs. More recently, PANDA \cite{panda} proposed expanding the feature dimensions of high-centrality nodes. However, this approach increases model parameters and discards potentially valuable information when selecting top-ranked features.

\textbf{Motivation.} 
To combat oversquashing, we go beyond the existing approaches like edge rewiring, increasing feature dimensions, or applying Transformers. The key reason behind oversquashing is accommodating the vast amount of information into fixed-dimensional feature vectors. We attempt to overcome this constraint with asynchronous feature updates. Despite storing the exponentially growing aggregated features simultaneously, we update the node features in a batch-wise manner. The grouping of the nodes never overflows the fixed-capacity feature vectors, effectively utilizing the same capacity. Recently, GwAC \cite{gwac} processes messages from nodes to their neighboring nodes at different times. Yet, this method overlooks the channel capacity and offers no provision to tackle oversquashing. On the contrary, our approach specifically proposes asynchronous message passing to confront oversquashing. 

\textbf{In this work}, we propose a novel plug-and-play framework \textbf{C}entrality-aware \textbf{A}synchronous \textbf{M}essage \textbf{P}assing or \textbf{CAMP} that adopts node centrality (degree, betweenness, etc) as the selection criterion to create node batches. The node centrality values are ordered (ascending or descending), and the entire set is partitioned into disjoint subsets. The number of subsets is fixed to the number of message-passing layers of the underlying GNN, and each subset is assigned to each layer. This partitioning ensures that every node gets a chance to exchange messages with its neighbors at some intermediate layer. The complete workflow is illustrated in Figure \ref{fig:camp_model}. In a network, the high-centrality nodes are assumed to be potential bottlenecks for message propagation. By processing these critical nodes earlier, CAMP allows their updated representations to flow to low-centrality nodes in subsequent layers. This approach strategically manages the information flow through these bottleneck nodes asynchronously. This mode of asynchronous updates prevents the simultaneous compression of higher- and lower-centrality node features into fixed-capacity channels. We also offer theoretical analyses to validate the efficacy of our proposed framework. 

\noindent
\textbf{Contribution.} Our contributions are succinctly outlined as: 
\begin{itemize}
    \item \textbf{(1) Asynchronous Message Passing:} we design a generalized framework for performing asynchronous message passing that departs from the conventional synchronous updates in standard GNNs like GCN \cite{gcn}, GAT \cite{gat}, GCNII \cite{gcnii}, etc. The framework selects node batches at each layer based on pre-defined criteria and updates only those features, with CAMP specifically using node centrality measures to guide batch selection. Our approach seamlessly integrates with diverse GNN architectures as a model-agnostic solution.

    \item \textbf{(2) Oversquashing Mitigation} Asynchronous message updates enable the processing of larger volumes of information through fixed-capacity channels at different layers. This prevents the simultaneous compression of exponentially growing information into fixed-sized vectors, effectively addressing the oversquashing without increasing model parameters or modifying graph structure. 

    \item \textbf{(3) Theoretical Insights} We provide extensive theoretical underpinnings demonstrating that CAMP mitigates oversquashing by enhancing feature sensitivity through its asynchronous updates.     
\end{itemize}

\section{Related Works}
The long-standing problem of Oversquashing undermines the prowess of GNNs, which is evident in tasks that require multi-hop interactions. Oversquashing was first observed by \cite{oversquashing} in the ambit of message-passing of GNNs. Several research attempts were made to navigate the challenging terrain of oversquashing, such as \cite{oversquashing_wdt,oversquashing_power}. Amidst all, rewiring emerged as the most enticing solution to tackle oversquashing. Earlier works like DropEdge \cite{dropedge} and GDC \cite{gdc} perform rewiring operations to enable long-range information exchange. A different class of methods emerged with SDRF \cite{sdrf}, GTR \cite{gtr}, BORF \cite{borf}, and SJLR \cite{sjlr} that characterize the exploration of edge-based curvatures. Those methods are non-differentiable and pursue sequential edge adjustments. The domain of spectral rewiring is further enriched with methods like FoSR \cite{fosr} to introduce relational rewiring, distinguishing between added and existing edges while optimizing the spectral gap of the network. In a different vein, \cite{banerjee_spectral} combines the information-theoretic properties with spectral expansion to enable efficient rewiring. A newer direction, unraveled with DiffWire \cite{diffwire} proposes an inductive differentiable rewiring strategy based on commute time, spectral gap, and effective resistance. The recently proposed LASER \cite{laser} sequentially rewires the graph, improving spatial connectivity and respecting the sparsity and locality of the input graph.
Additionally, DRew \cite{drew} dynamically rewires the graph with a fixed delay in interactions among the neighbors. Another work, Co-GNN \cite{cognn} performs layer-wise dynamic rewiring that learns when to update node features, highlighting asynchronous properties. A similar category of work \cite{async_cikm} proposed aggregating information and updates after some fixed delay, which also satisfies the asynchronous property. Another effective way to rewire is by converting the input into another graph by preserving the expander-type properties \cite{egp}. PR-MPNN \cite{prmpnn} devised a rewiring strategy to construct graph structure depending on the downstream tasks and underlying data distribution. The most recent work, PANDA \cite{panda} enhances the channel capacity of the nodes with higher importance to process a larger volume of information effectively to prevent oversquashing. Another line of work, GOKU \cite{spectrum_preserving} rewires the graph topology by preserving the spectrum of the input graph. \cite{short_range_oversq} studies the presence of oversquashing even in the short-range tasks.  

Recently, the effects of Oversquashing have been investigated beyond static graphs. \cite{oversquashing_tempo} reformulated the sensitivity bound of oversquashing for temporal graphs. Furthermore, the impact of oversquashing is studied for hypergraphs \cite{oversquashing_hypergraph}. On a different vein, \cite{oversquashing_dynamical} understands the problem of oversquashing assisted by the concept of dynamical systems.

\section{Proposed Method}

\subsection{Preliminaries \& Notations}
Assume an attributed graph $\mathcal{G}=(\mathcal{V}, \mathcal{E}, X, A)$ where $\mathcal{V}$ denotes the set of nodes, $\mathcal{E} \subseteq \mathcal{V} \times \mathcal{V}$ denotes the edge set, $X \in \mathbb{R}^{n \times d}$ is the feature matrix containing $d$-dimensional node features, and $A \in \mathbb{R}^{n \times n}$ is the adjacency matrix. The neighborhood of a node $v$ is represented as $N(v)$.  

\noindent
\textbf{Revisiting MP-GNN.}
The existing MP-GNN computes messages between the pair of adjacent nodes $(u, v)$ by a \texttt{MESSAGE} function $\Psi$, and subsequently messages are aggregated across the neighborhood $N(v)$. The features of the centering node $v$ are updated by a \texttt{COMBINE} function $\Phi$. Notably, the feature updates are performed simultaneously, and thereby MP-GNN performs "synchronous" message passing across neighborhoods. The updated features of node $v$ is $x_{v}^{(l+1)}$ can be represented as:   
\begin{equation}
x_{v}^{(l+1)} = \Phi_{l} \left (x_{v}^{(l)} , \sum\limits_{u \in N(v)} \Psi_{l}(x_{v}^{(l)}, x_{u}^{(l)}) \right)    
\end{equation}
where $\Psi_{l}$ and $\Psi_{l}$ ar respectively layer-specific message and combine function.  

\noindent
\textbf{Oversmoothing.} 
The problem mostly persists in deep MP-GNNs, causing performance degradation when long-range interaction is required \cite{oono2019graph, oversmoothing_notes}. As a result, node features become indistinguishable, influenced by the recursive neighborhood aggregation in the deeper layers. The growing receptive field in deep layers became almost identical and thereby computes similar embeddings for each node, resulting in oversmoothing. In this work, we will demonstrate that asynchronous message passing may alleviate oversmoothing in deep GNNs.

\noindent
\textbf{Oversquashing.}
The phenomenon \cite{oversquashing} occurs when compressing aggregated features from a large number of neighbors into fixed-capacity channels. The effect of oversquashing can be measured by the sensitivity bounds proposed by \cite{oversquashing_wdt, oversquash_ricciflow}. assuming there exists a $l$-length path between $u$ and $v$. The sensitivity bound is estimated as the Jacobian of $h^{(l)}_{v}$ and $h^{(0)}_{u}$ which can be represented as,
\begin{equation}
\label{eq:camp_oversq_eq}
     \norm{\frac{\partial h^{(l)}_{v}}{\partial h^{(0)}_{u}}}_{1} \le \underbrace{(cwp)^{l}}_{\text{model}} \underbrace{(\Tilde{S}^{l})_{uv}}_{\text{topology}},
\end{equation}
where $c$ is the Lipschitz constant of the model parameters, $w$ is the maximum entry of the weight matrices, and $p$ is the width. Also, oversquashing is affected by $\Tilde{S}^{l}$, where $\Tilde{S}$ can be a normalized adjacency matrix. The lower value of the bound signals a diminished effect on the message propagation of a node by its neighbors.



\subsection{Asynchronous Message Passing Framework}
In this section, we lay the foundation of the generalized framework for asynchronous feature updates in GNNs. Given a pre-defined node selection strategy $\mathcal{S}$, a batch of nodes, say $\text{B}$, with $|\text{B}| \le n$ is chosen. The features of nodes in $\text{B}$ will only be updated by neighborhood aggregation, and the features of the remaining nodes will be retained for the next layer. Propagation through hidden layers can be perceived as an increase in layers and also an increase in the timestamp. If the feature of a node $v \in B$ is updated at $l^{th}$ layer, then it also crosses the time-stamp $l$. We tag every node with a tuple representing the layer number and the corresponding time-stamp. The following equation represents the asynchronous message-passing operation conditioned on $\mathcal{S}$,   
\begin{equation}
\label{eq:amp_update}
   x_{v}^{(l+1, l+1)} = 
   \begin{cases}
       & \Phi_{l} \left (x_{v}^{(l_v, l)} , \sum\limits_{u \in N(v)} \Psi_{l}(x_{v}^{(l_v, l)}, x_{u}^{(l_u, l)}) \right), \\ & \hspace{3cm} \forall  v \in \text{B} | \mathcal{S} \\
        & x_{v}^{(l_v, l+1)}, \,\, \forall v \not\in \text{B} | \mathcal{S}.
   \end{cases}  
\end{equation}
where $\Psi_l, \Phi_l$ are respectively \texttt{MESSAGE} and \texttt{COMBINE} functions for $l^{th}$ layer, $l_v$ denotes the layer at which $x_v$ is last updated with $0 \le l_v \le l$. If a node does not participate in the message-passing operation, then its time-stamp will increase by one, but the last updated layer remains unchanged. Importantly, the framework considers a node set which not necessarily need to be part of any subgraph, strictly differing from the methods pursuing subgraph sampling like \cite{graphsage, graphsaint} 


\subsection{Centrality-aware Asynchronous Message Passing}
We consider node centrality values as the selection criteria $\mathcal{S}$, prioritizing the importance of the nodes in the given network. We adopt five well-recognized centrality measures: Degree \cite{degree}, Betweenness \cite{betweenness}, Closeness \cite{closeness}, Load \cite{load}, and PageRank \cite{pagerank}. Now we have the idea of 'which' nodes to be chosen but not the idea of 'when'. The problem is resolved by assigning each node to a particular layer. 
 
Initially, we estimate the centrality scores of each node in the pre-processing stage. Concurrently, we also assign an ordering (either ascending or descending) of the centrality values, which confirms the monotone effects of nodes' importance and discards random choices. For instance, the centrality of nodes in descending order can be presented as follows,
\begin{equation}
    C_n = \{c^{(1)}, c^{(2)}, \cdots, c^{(n)}: c^{(i+1)} \ge c^{(i)}, \forall \,\, 1 \le i \le n-1\},
\end{equation}
where $c^{(i)}$ denotes the centrality of the node $i$. To ensure the participation of every node at some layer, we partition the centrality values into $L$ smaller subsets, where $L$ denotes the total number of layers in the GNN. Note that each subset maintains a similar order to $C_n$. The $l^{th}$ subset can be presented as,
\begin{equation}
    C^{(l)}_n = \{c^{(1)}_l, c^{(2)}_l, \cdots, c^{(m)}_l: c_{l}^{(i+1)} \ge c_{l}^{(i)}, \forall \,\, 1 \le i \le m-1\}
\end{equation}
where $c^{(i)}_{k}$ is the centrality of $i^{th}$ node which belong to the $l^{th}$ subset and $m = \lvert C^{(l)}_n \rvert = \lfloor \frac{C_n}{L}  \rfloor$ with $l \in [1, L]$. 

\noindent
\textbf{Node-batch Selection} 
We select a batch of nodes at every layer to update features. For layer $l$, we consider the subset $C_{n}^{l}$ as our candidate node batch. The disjoint partition of $C_n$ guarantees the layer-dependent participation of nodes during a forward pass. We also offer a provision for selecting a portion of nodes from each subset $C_{n}^{l}$. Assuming a sampling probability is $p$, then the size of the node batch will be $mp = \lvert C_{n}^{l} \rvert p$, where $0 < p \le 1$. We term the framework as $p$-CAMP. If we select a full batch at every layer, then our framework is termed $1$-CAMP or simply CAMP.  

\noindent
\textbf{Update Rule} The feature update rule of CAMP can be defined as follows,
\begin{equation}
\label{eq:camp_update}
   x_{v}^{(l+1, l+1)} = 
   \begin{cases}
       & \Phi_{l} \left (x_{v}^{(l_v, l)} , \sum\limits_{u \in N(v)} \Psi_{l}(x_{v}^{(l_v, l)}, x_{u}^{(l_u, l)}) \right), \\ & \hspace{3cm} \forall  v \in C^{(l)}_{n} \\
        & x_{v}^{(l_v, l+1)}, \,\, \forall v \not\in C^{(l)}_{n},
   \end{cases}  
\end{equation}
where all notations carry similar meanings as mentioned in Eq~\ref{eq:amp_update}. 

The features of a candidate node batch get updated by aggregating features from their immediate neighbors. The current updates rely on the older versions of the neighboring features. Furthermore, the updated features may remain unchanged for a series of hidden layers and can be utilized by other node batches in subsequent layers. The layer-dependent access prevents the simultaneous usage of fixed-capacity channels and guarantees efficient resource utilization. In the following section, we will discuss the theoretical justifications of the effectiveness of CAMP.

\subsection{Theoretical Analysis}
In this section, we provide theoretical underpinnings for CAMP in mitigating oversquashing. 

\begin{definition}[Valid Asynchronous Walk]
    A walk $\pi = (i_0 = s,\, i_1,\, \ldots,\, i_K = v)$ of length $K$
    from $s$ to $v$ is a \emph{valid asynchronous walk} under the
    schedule $\{C_n^{(\ell)}\}_{\ell=1}^{L}$ if there exist strictly
    increasing layer indices $\tau_1 < \tau_2 < \cdots < \tau_K$ with
    $\tau_t \in [L]$ such that
    \begin{enumerate}
        \item $(i_{t-1}, i_t) \in \mathcal{E}$ for every $t \in [K]$
              (each step follows an edge), and
        \item $i_t \in C_n^{(\tau_t)}$ for every $t \in [K]$
              (node $i_t$ is active at layer $\tau_t$, the layer at
               which the edge $(i_{t-1} \to i_t)$ is traversed).
    \end{enumerate}
    We write $\mathcal{P}_{\mathrm{async}}(s,v)$ for the set of all valid
    asynchronous walks from $s$ to $v$, and $K(\pi)$ for the length of
    walk $\pi$. Define
    \[
        K(s,v) \;:=\; \min_{\pi \in \mathcal{P}_{\mathrm{async}}(s,v)}
        K(\pi).
    \]
    Since the partition has $L$ parts and at most one edge traversal can
    occur per layer, $K(s,v) \leq \min\!\left(L,\,
    d_G(s,v)\right)$, where $d_G(s,v)$ is the
    shortest-path distance in $G$.
\end{definition}

\begin{lemma}[Asynchronous Sensitivity Bound for CAMP]
\label{lemma:camp-sensitivity}
Consider an $L$-layered GNN applied to a graph $G = (V, E)$ integrated with the CAMP framework under activation schedule
$\{C_n^{(\ell)}\}_{\ell=1}^{L}$. Assume that for every layer $\ell$, the Jacobians of $\Phi_\ell$ and $\Psi_\ell$ satisfy the Lipschitz bounds
\begin{equation}
    \left\|\nabla\Phi_\ell\right\| \leq \alpha, \qquad
    \left\|\nabla\Psi_\ell\right\| \leq \beta,
    \label{eq:lip}
\end{equation}
for real constants $\alpha, \beta > 0$.

\noindent\emph{(Part I: Sensitivity Bound.)}
For any pair of nodes $s, v \in V$,
\begin{equation}
    \left\|\frac{\partial x_v^{(L)}}{\partial x_s}\right\|_1
    \;\leq\;
    (\alpha\beta)^{K(s,v)}
    \sum_{\pi \in \mathcal{P}_{\mathrm{async}}(s,v)}
    \prod_{t=1}^{K(\pi)} S^{(\tau_t)}_{i_t,\, i_{t-1}}.
    \label{eq:camp_bound}
\end{equation}

\noindent\emph{(Part~II: Depth-wise Dominance over Synchronous
MP.)}
The synchronous sensitivity bound satisfies
\begin{equation}
    \left\|\frac{\partial x_v^{(L)}}{\partial x_s}
    \right\|_1^{\mathrm{sync}}
    \;\leq\;
    (\alpha\beta)^{L}\,(S^L)_{vs}.
    \label{eq:sync_bound}
\end{equation}
If $0 < \alpha\beta \leq 1$ and there exists at least one valid
asynchronous walk $\pi^* \in \mathcal{P}_{\mathrm{async}}(s,v)$ with
$K(\pi^*) = K(s,v) < L$, then
\begin{equation}
    \frac{\text{CAMP bound in \eqref{eq:camp_bound}}}
         {\text{Sync bound in \eqref{eq:sync_bound}}}
    \;\geq\;
    (\alpha\beta)^{K(s,v) - L}
    \;\xrightarrow{L\to\infty}\; +\infty,
    \label{eq:dominance}
\end{equation}
with $K(s,v)$ held fixed. In particular, for fixed graph distance
$d_G(s,v) = D$ and increasing network depth $L$, the synchronous
bound vanishes exponentially in $L$ while the CAMP bound's prefactor
$(\alpha\beta)^{K(s,v)}$ remains bounded below by $(\alpha\beta)^{D}$,
independent of $L$.
\end{lemma}

\begin{remark}
Equation~\eqref{eq:dominance} does \emph{not} claim that CAMP's bound is universally larger than the synchronous bound. The path-sum $\sum_\pi \prod_t S^{(\tau_t)}$ is generally smaller than $(S^L)_{vs}$ because $\mathcal{P}_{\mathrm{async}}(s,v) \subseteq \mathcal{P}_{\mathrm{sync}}(s,v)$. The correct interpretation is: CAMP decouples the contraction depth from the network depth, which is the exponent governing exponential decay, $K(s,v) \leq D$, not $L$. This is the precise mechanism by which CAMP alleviates oversquashing in deep networks.
\end{remark}

\begin{figure}[!ht]
    \centering
    \includegraphics[width=0.90\textwidth]{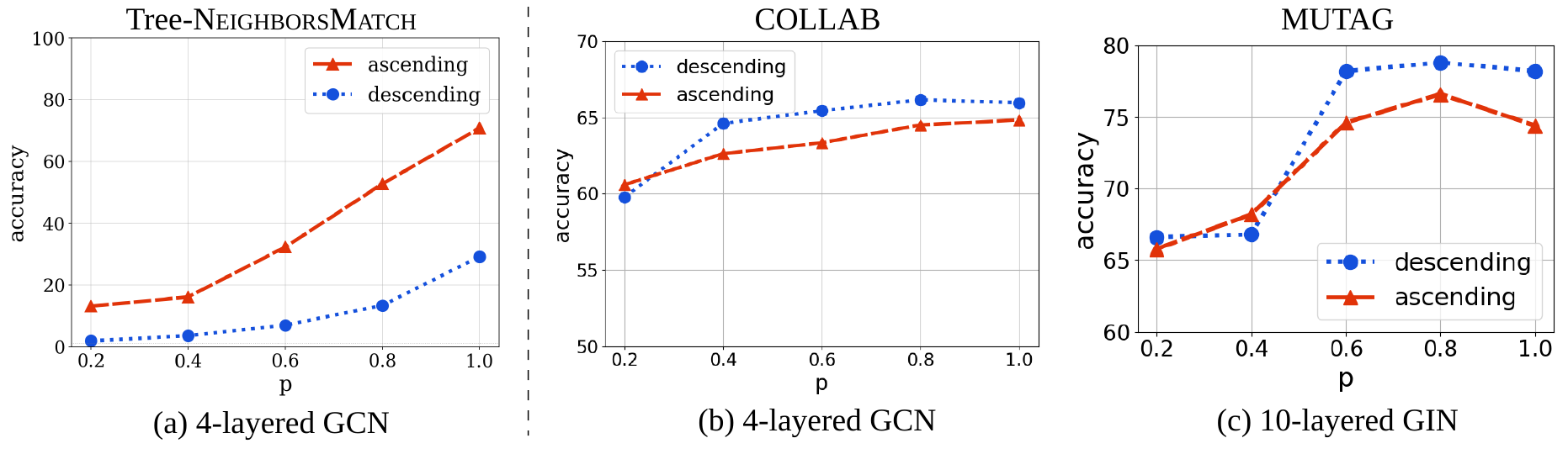}
    \caption{The comparative study between the order of node centrality values and the performance of CAMP is presented. The performance is evaluated on the various batch sampling rates. }
    \label{fig:order_main}
\end{figure}

\begin{table*}[!ht]
\centering
\scriptsize
\caption{The performances of GCN and GIN coupled with CAMP are presented. For each dataset, the best and second-best results are respectively marked in {\color{Green}green} and {\color{Blue}blue} colors. }
\label{tab:main_table}
\resizebox{0.90\columnwidth}{!}{
\begin{tabular}{l|cccccc}
\toprule
Method & \textsc{Enzymes} & \textsc{Mutag} & \textsc{Proteins} & \textsc{Collab} & \textsc{IMDB-Binary} & \textsc{Reddit-Binary} \\
\midrule
GCN (None) & $27.66 \pm 1.16$ & $72.15 \pm 2.44$ & $70.98 \pm 0.73$ & $33.78 \pm 0.48$ & $49.77 \pm 0.81$ & $68.25 \pm 1.09$ \\
+ Last Layer FA & $26.46 \pm 1.20$ & $70.05 \pm 2.02$ & $71.01 \pm 0.96$ & $33.32 \pm 0.43$ & $48.98 \pm 0.94$ & $68.48 \pm 0.94$ \\
+ Every Layer FA & $18.33 \pm 1.03$ & $70.45 \pm 1.96$ & $60.03 \pm 0.92$ & $51.79 \pm 0.41$ & $48.17 \pm 0.80$ & $48.49 \pm 1.04$ \\
+ DIGL & $27.51 \pm 1.05$ & $71.35 \pm 2.39$ & $70.60 \pm 0.73$ & $53.35 \pm 0.29$ & $49.91 \pm 0.84$ & $49.98 \pm 0.68$ \\
+ SDRF & $28.36 \pm 1.17$ & $71.05 \pm 1.87$ & $70.92 \pm 0.79$ & $33.44 \pm 0.47$ & $49.40 \pm 0.90$ & $68.62 \pm 0.85$ \\
+ FoSR & $25.06 \pm 0.99$ & $80.00 \pm 1.57$ & $73.42 \pm 0.81$ & $33.83 \pm 0.58$ & $49.66 \pm 0.86$ & $70.33 \pm 0.72$ \\
+ BORF & $24.70 \pm 1.00$ & $75.80 \pm 1.90$ & $71.00 \pm 0.80$ & Time-out & $50.10 \pm 0.90$ & Time-out \\
+ GTR & $27.52 \pm 0.99$ & $79.10 \pm 1.86$ & $72.59 \pm 2.48$ & $33.05 \pm 0.40$ & $49.92 \pm 0.99$ & $68.99 \pm 0.61$ \\
+ CT-Layer & $17.38 \pm 1.03$ & $75.89 \pm 3.02$ & $60.35 \pm 1.06$ & $52.14 \pm 0.41$ & $50.32 \pm 0.94$ & $51.58 \pm 1.01$ \\
+ PANDA & \color{Blue} $\mathbf{31.55 \pm 1.23}$ & \color{Green} $\mathbf{85.75 \pm 1.39}$ & \color{Green} $\mathbf{76.00 \pm 0.77}$ & \color{Blue} $\mathbf{68.40 \pm 0.45}$ & $\color{Green} \mathbf{63.76 \pm 1.01}$ & \color{Blue} $\mathbf{80.69 \pm 0.72}$ \\
+ GOKU & $27.60\pm1.20$ & $81.00\pm2.00$ & $71.90\pm0.80$ & - & - & - \\
\midrule
+ CAMP (Ours) & \color{Green} $\mathbf{31.73\pm2.30}$ & $\color{Blue} \mathbf{81.80\pm3.68}$ & \color{Blue} $\mathbf{74.43\pm1.64}$ & \color{Green} $\mathbf{69.86\pm0.70}$ & \color{Blue} $\mathbf{62.44\pm2.15}$ & $\color{Green} \mathbf{84.96\pm0.86}$ \\
\midrule
GIN (None) & $33.80 \pm 0.11$ & $77.70 \pm 0.36$ & $70.80 \pm 0.82$ & $72.99 \pm 0.38$ & $70.18 \pm 0.99$ & $86.78 \pm 1.05$ \\
+ Last Layer FA & $44.40 \pm 1.38$ & $83.05 \pm 1.74$ & $72.30 \pm 0.66$ & $71.05 \pm 0.40$ & $70.91 \pm 0.78$ & $88.31 \pm 0.65$ \\
+ Every Layer FA & $28.38 \pm 1.05$ & $72.55 \pm 3.01$ & $70.37 \pm 0.91$ & $32.98 \pm 0.39$ & $49.16 \pm 0.87$ & $50.36 \pm 0.68$ \\
+ DIGL & $35.71 \pm 1.19$ & $79.70 \pm 2.15$ & $70.75 \pm 0.77$ & $54.50 \pm 0.41$ & $64.39 \pm 0.90$ & $76.03 \pm 0.77$ \\
+ SDRF & $35.81 \pm 1.00$ & $78.40 \pm 2.80$ & $69.81 \pm 0.79$ & $72.95 \pm 0.41$ & $69.72 \pm 1.15$ & $86.44 \pm 0.59$ \\
+ FoSR & $29.20 \pm 1.36$ & $78.40 \pm 2.80$ & $73.10 \pm 0.81$ & $73.27 \pm 0.41$ & $71.21 \pm 0.91$ & $87.35 \pm 0.59$ \\
+ BORF & $35.50 \pm 1.20$ & $80.80 \pm 2.50$ & $73.70 \pm 0.80$ & Time-out & $71.30 \pm 1.50$ & Time-out \\
+ GTR & $30.57 \pm 1.42$ & $77.60 \pm 2.84$ & $73.13 \pm 0.69$ & $72.93 \pm 0.42$ & $71.28 \pm 0.86$ & $86.98 \pm 0.66$ \\
+ CT-Layer & $16.58 \pm 0.90$ & $56.85 \pm 4.25$ & $61.10 \pm 1.18$ & $52.30 \pm 0.60$ & $50.00 \pm 0.97$ & $54.58 \pm 1.75$ \\
+ PANDA & \color{Blue} $\mathbf{46.20 \pm 1.41}$ & $\color{Green} \mathbf{88.75 \pm 1.57}$ & \color{Blue} $\mathbf{75.75 \pm 0.85}$ & \color{Green} $\mathbf{75.11 \pm 0.21}$ & \color{Blue} $\mathbf{72.56 \pm 0.91}$ & $\color{Green} \mathbf{91.05 \pm 0.40}$ \\
+ GOKU & $33.80\pm1.20$ & $78.40\pm2.50$ & $73.90\pm1.00$ & - & - & - \\
\midrule
+ CAMP (Ours) & \color{Green} $\mathbf{46.60\pm2.26}$ & \color{Blue} $\mathbf{87.40 \pm3.95}$ & \color{Green} $\mathbf{76.07\pm1.73}$ & \color{Blue} $\mathbf{74.14\pm0.65}$ & \color{Green} $\mathbf{73.48\pm1.47}$ & \color{Blue} $\mathbf{89.24\pm1.03}$ \\
\bottomrule
\end{tabular}}
\end{table*}

\subsection{Properties of CAMP}

\textbf{[1] Does order (ascending or descending) matter?}
The effectiveness of CAMP can be influenced by the ordering of node centrality values. We consider two types of datasets: (1) Tree-\textsc{Neighbors}\textsc{Match} \cite{oversquashing} and (2) COLLAB and MUTAG to examine this effect. Figure~\ref{fig:order_main} reports the results. For the first dataset, CAMP performs better with centrality values sorted in ascending order, whereas for the second set, it performs better with a descending order. NeighborsMatch dataset is a binary tree with directed edges exists only from child to parent. Thus, the information will always flow from leaf nodes to the root. If we apply descending ordering, the root node will get activated at the initial layer. It will aggregate from the neighbors that are not updated by the GNN at all. In the subsequent layer, the child nodes will get updated, but the information will never reach the root node. In contrast, the ascending ordering will first updated the features of the child nodes, and then the updated embeddings will flow to the root node. 

On a different note, COLLAB or MUTAG has no such single direction of information flow. The graphs of these datasets contain hub nodes that relay to the other nodes. Activating high-centrality (bottleneck) nodes early allows their enriched representation to propagate broadly to the low-centrality nodes. Furthermore, nodes with high centrality connect a larger volume of edges, causing oversmoothing in deeper layers. For a descending order of $C_{n}$, the faster interaction of high-centrality nodes compared to lower ones is supposed to diminish the effect of oversmoothing.


\noindent
\textbf{[2] Variable Hop Aggregation.}
CAMP simulates a variable hop aggregation strategy, unlike the standard MP-GNNs. For any layer $l$, the corresponding candidate node batch $C_{l}^{n}$ updates features from the neighbors, which can be either already updated in some previous layers $(<l)$ or will be updated in the upcoming layers $(>l)$. The updated features of the current node batch are the aggregated form of variable hops and the updated features of their corresponding neighbors. This variable hop aggregation is the key characteristic of the asynchronous message-passing frameworks.  


\begin{table}[!ht]
\centering
\scriptsize
\caption{Results of GCN and GIN with None, SDRF, FoSR, BORF, PANDA, and CAMP (Ours) on \textsc{Peptides-func} and \textsc{Peptides-struct}. The best and second-best results are respectively marked in {\color{Green}green} and {\color{Blue}blue}.}
\label{tab:peptides_results}
\begin{tabular}{l|cccc}
\toprule
\multirow{2}{*}{Method} & \multicolumn{2}{c}{\textsc{Peptides-func} (AP $\uparrow$)} & \multicolumn{2}{c}{\textsc{Peptides-struct} (MAE $\downarrow$)} \\
\cmidrule{2-5}
 & GCN & GIN & GCN & GIN \\
\midrule
None & $59.30 \pm 0.23$ & \color{Blue}$\mathbf{64.17\pm0.0046}$ & $0.3496 \pm 0.0013$ & $33.24\pm0.0041$ \\
+ SDRF & $59.47 \pm 1.26$ & \color{Green}$\mathbf{66.86\pm0.0005}$ & $0.3478\pm 0.0013$ & $32.92\pm0.0008$ \\
+ FoSR & $59.47 \pm 0.35$ & $63.72\pm0.0031$ & $0.3473 \pm 0.0007$ & $34.73\pm0.0042$ \\
+ BORF & \color{Blue} $\mathbf{59.94 \pm 0.37}$ & $57.61\pm0.0001$ & $0.3514 \pm 0.0009$ & $34.89\pm0.0006$ \\
+ PANDA & \color{Green} $\mathbf{60.28 \pm 0.31}$ & $59.03\pm0.0071$ & \color{Blue}$\mathbf{0.3272 \pm 0.0001}$ & $35.62\pm0.0031$ \\
\midrule
+ CAMP (Ours) & $56.67 \pm 0.76$ & $54.83\pm0.0023$ & \color{Green} $\mathbf{0.3138 \pm 0.0041}$ & \color{Green}$\mathbf{37.92\pm0.0061}$ \\
\bottomrule
\end{tabular}
\end{table}

\subsection{Complexity Analysis}
The time complexity of CAMP-GNN depends solely on the underlying GNN model, since the CAMP framework adds only an extra task: computing centrality scores, which is performed as a pre-processing step. The cost of computing centrality values depends on the choice of centrality; the cost of degree centrality is $\mathcal{O}(|\mathcal{E}|)$, the same for closeness is $\mathcal{O}(|\mathcal{V}||\mathcal{E}|)$, etc. Unlike PANDA, we do not increase the number of feature dimensions during message propagation; thus, the parameter complexity for CAMP-GNN also remains unaffected. Note that estimating node centrality can be computationally expensive for large-scale graphs. Thus, we bound our experiments to molecular graphs, which is also our key limitation.

\begin{table*}[!ht]
\centering
\scriptsize
\caption{A comparative study of the performance of CAMP-GCN based on five centrality measures. The best results are boldfaced.  }
\label{tab:centrality_gcn}
\resizebox{0.90\columnwidth}{!}{
\begin{tabular}{lccccccc}
\toprule
Centrality & \textsc{ENZYMES} & \textsc{MUTAG} & \textsc{PROTEINS} & \textsc{COLLAB} & \textsc{IMDB-BINARY} & \textsc{REDDIT-BINARY} \\
\midrule
Degree & $28.93\pm2.65$ & $78.40\pm4.46$ & $72.00\pm1.75$ & $69.25\pm0.71$ & $58.52\pm2.34$ & $80.10\pm0.90$ \\
Betweenness & $30.20\pm2.56$ & $78.20\pm4.20$ & $71.54\pm1.58$ & $69.81\pm0.82$ & $58.24\pm2.32$ & $79.90\pm0.94$ \\
PageRank & $29.40\pm2.37$ & $\mathbf{78.40\pm4.07}$ & $71.14\pm1.90$ & $69.75\pm0.86$ & $\mathbf{60.48\pm1.89}$ & $81.28\pm1.14$ \\
Load & $31.13\pm2.22$ & $78.20\pm4.16$ & $72.11\pm1.98$ & $\mathbf{69.86\pm0.70}$ & $59.32\pm1.95$ & $79.94\pm1.06$ \\
Closeness & $\mathbf{31.73\pm2.30}$ & $77.60\pm3.88$ & $\mathbf{74.43\pm1.64}$ & $69.25\pm0.83$ & $59.52\pm2.14$ & $\mathbf{84.96\pm0.86}$ \\
\bottomrule
\end{tabular}}
\end{table*}

\subsection{Instances of CAMP}
We will now illustrate the examples of CAMP-GNN by using GCN \cite{gcn} and GIN \cite{gin}. The layer-wise update rule of CAMP-GCN is,  
\begin{equation}
\label{eq:camp_gcn_framework}
    h_{u}^{(l+1, l+1)} = h_{u}^{(l, l)} + \sigma \Big( \hspace{-0.75cm} \sum\limits_{v \in N(v) \cap C^{(l+1)}_{n}} \hspace{-0.5cm} W^{(l+1)} h_{v}^{(l_v, l)}  \Big ), 
\end{equation}
where $W^{(l+1)}$ is the parameterized weight matrix. Observe that neighborhood aggregation is performed on the nodes that are neighbors and belong to the candidate node batch. Similarly, we define the update rule of CAMP-GIN as follows:
\begin{equation}
\label{eq:camp_gin_framework}
    h_{u}^{(l+1, l+1)} = \text{MLP}_{l} \Big( (1+\epsilon) h_{u}^{(l, l)} + \hspace{-0.75cm} \sum\limits_{v \in N(v) \cap C^{(l+1)}_{n}} \hspace{-0.5cm} W^{(l+1)} h_{v}^{(l_v, l)}  \Big),
\end{equation}
where $\text{MLP}_l$ denotes the combination of the linear layers with ReLU activation and $\epsilon$ is the learnable parameter. In this case, the node features are also considered for aggregation if they belong to both the neighborhood and the candidate batch.  

\vspace{-10pt}
\section{Experiments}

\noindent
\textbf{Datasets.}
For graph classification, we considered six benchmark datasets, ENZYMES, MUTAG, PROTEINS, COLLAB, IMDB-BINARY, and REDDIT-BINARY, obtained from TUDatasets \cite{tudataset}. Additionally, we included Peptides-func and Peptides-struct from LRGB datasets \cite{lrgb} which require long-range interactions for expected performance. The details of the datasets are discussed in Section \ref{camp_data_details} of the Appendix. 

\subsection{Experimental Settings.}
We applied CAMP coupled with GCN and GIN to solve downstream tasks. We compare performance with the no-rewiring method and several baselines: Fully Adjacent (FA), DIGL, SDRF, FoSR, BORF, GTR, CT-layer, PANDA, and GOKU. We split each dataset into $80/10\%/10\%$ for the train. validation, and test, respectively. For TUDatasets, we executed experiments for $25$ randomly generated splits and produced mean and standard deviations over the test set, and for LRGB, we used the standard splits. More details on the experimental setup and hyperparameters are provided, respectively, in Section \ref{camp_Exp_setup} and Section \ref{camp_hyper_details} of the Appendix. Our implementation is available at \href{https://github.com/kushalbose92/camp}{https://github.com/kushalbose92/camp}.   

\subsection{Results \& Discussions.}
The performance of CAMP paired with GCN and GIN is presented in Table \ref{tab:main_table}, suggesting our proposed framework either outperforms all contenders or attains a second position, outperforming at least all rewiring methods. 
CAMP showed commendable improvements on REDDIT-BINARY, which contains larger graphs with multiple bottleneck nodes. CAMP successfully enables information propagation, eventually mitigating oversquashing. CAMP also outperformed on COLLAB, which contains a scientific collaboration network with a community structure. Our framework efficiently identifies key bottleneck nodes and enables message flow asynchronously to overcome channel constraints. CAMP is ranked as the runner-up on MUTAG, PROTEINS, and IMDB-BINARY due to its smaller size, high sparsity, and fewer bottleneck nodes. The asynchronous updates convert most of the nodes into an isolated state that plagues effective message passing. 

The performance on Peptides datasets is presented in Table \ref{tab:peptides_results}. CAMP improves Peptides-struct but underperforms the baseline on Peptides-func despite both datasets sharing identical graphs and node features, differing only in the prediction task. This asymmetry arises because Peptides-struct's targets are global geometric properties that aggregate (e.g., sphericity, inertia tensor) that benefit from the asynchronous updates. In contrast, labels of Peptides-func depend on local motifs, so CAMP's delayed updates only cost accuracy, with no long-range gain to offset it. 

\begin{figure}[!ht]
    \centering
    \includegraphics[width=\linewidth]{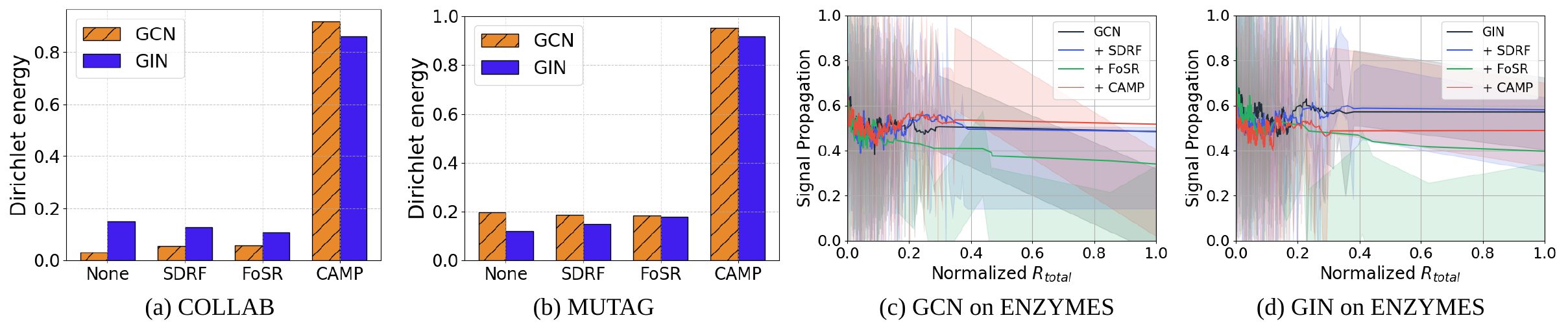}
    \caption{A comparative study on Dirichlet energies is presented in (a) and (b). CAMP performs better than other rewiring methods. The signal propagation with respect to normalized $R_{\text{total}}$ is presented in (c) and (d). CAMP maintains steady trends compared to others, with increasing resistance.  }
    \label{fig:energy_signal_combined}
\end{figure}


\subsection{Effect of Centrality Measure}
We studied the effect of various node centrality measures on the performance of CAMP. We employed $10$-layer GCN and GIN models as our backbones across six datasets, and the results are shown in Table \ref{tab:centrality_gcn}. For instance, CAMP-GCN performed best on PROTEINS and REDDIT-BINARY when closeness centrality was employed. In IMDB-BINARY, PageRank was the most suitable centrality measure for achieving optimal performance. The variation in performance underscored the utility of employing a diverse array of centrality measures. The results for CAMP-GIN are provided in Section \ref{camp_centrality_gin_subsection} of the Appendix.


\subsection{Variation of Dirichlet Energy}
We applied $10$-layered GCN and GIN paired with CAMP on COLLAB and MUTAG and estimated the Dirichlet energies for the final layers. Figure~\ref{fig:energy_signal_combined}(a) and \ref{fig:energy_signal_combined}(b) suggests Dirichlet energy for CAMP is higher than that of familiar rewiring methods like FoSR and SDRF. The node features remained distinguishable in deeper layers, confirmed by the higher energy values. Thus, CAMP improves the model performance in multi-layered GNNs without pursuing any graph rewiring. More results are available in Section \ref{camp_de_subection} of the Appendix. 

\subsection{Signal Propagation} 
The total effective resistance ($R_{\text{total}}$) \cite{siam_resistance} measures the obstacles to propagating information across the network. A graph with higher effective resistance is highly likely to be grappled with oversquashing \cite{gtr}. The efficiency of CAMP is measured by signal propagation relative to the normalized $R_{\text{total}}$ \cite{oversquashing_wdt}. Figures \ref{fig:energy_signal_combined}(c) and \ref{fig:energy_signal_combined}(d) illustrate that CAMP performs better compared to other rewiring methods by effectively propagating the signal across the network with increasing effective resistance. More results are available in Section \ref{camp_signal_prop} of the Appendix.   

\begin{figure}[!ht]
    \centering
    \includegraphics[width=\linewidth]{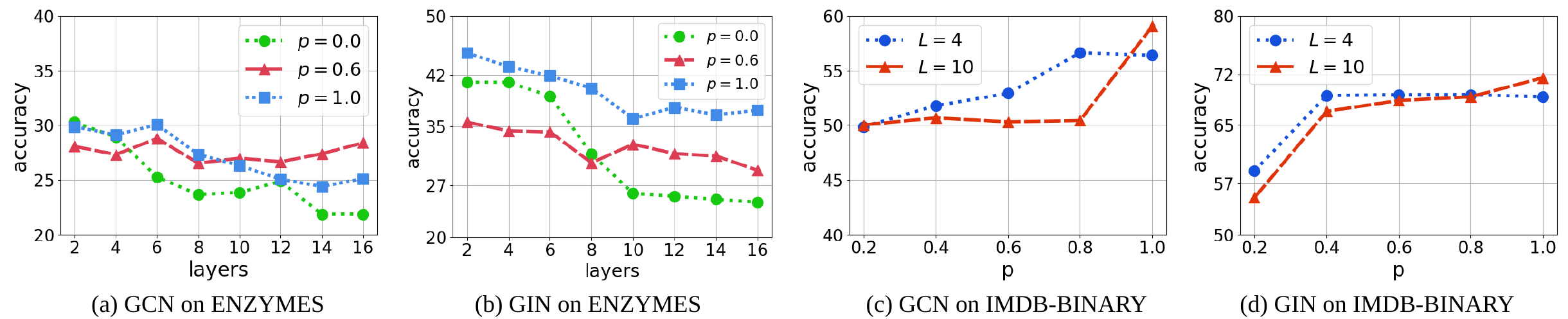}
    \caption{The performance of CAMP in deeper layers is presented in (a) and (b). The performance improved when the node batch sampling rate was increased, even as the number of layers increased. On the other hand, the performance for various node batch sampling rates for two different numbers of model layers is presented in (c) and (d). Performance improves with higher sampling rates. }
    \label{fig:batch_layers_combined}
\end{figure}

\subsection{Effect on Oversmoothing}
We studied the effect of oversmoothing in multi-layered GCN and GIN on ENZYMES at various node batch sampling rates. Figures~\ref{fig:batch_layers_combined}(a) and \ref{fig:batch_layers_combined}(b) delineate the performance trends in the various network depths with varying $p$. CAMP demonstrated resilience to oversmoothing for the higher values of $p$. The performance strictly degrades when CAMP is not applied, asserting the prowess of the proposed framework in tackling oversmoothing. The batching of the nodes prevents the simultaneous aggregation of the features from the growing receptive fields. Furthermore, the relative performance also improves when $p$ increases due to more participating nodes in each batch. The results for the remaining datasets are provided in Section \ref{camp_oversmooth_subsection} of the Appendix. 


\subsection{Variation of Batch Size}
We studied the impact of various node batch sampling rates $p$ on the performance of GCN and GIN. The experiment was conducted on IMDB-BINARY, and classification accuracy is presented in Figure~\ref{fig:batch_layers_combined}(c) and \ref{fig:batch_layers_combined}(d) for two different model depths. The trends illustrate that an increase in the sampling rate leads to an uptrend in performance. This is due to the larger number of nodes participating in the message-passing operation and aggregating more relevant information with the increasing $p$. The additional plots are presented in Section \ref{camp_batch_subsection} of the Appendix. 


\subsection{Limitations}
The centrality values for the nodes are computed in the pre-processing steps. Some centrality measures, like Betweenness, consumes an ample of amount of time. In this work, the impact of pre-processing time is moderate due to the involvement of small-sized molecular graphs (Peptides-func) or mid-sized social network graphs (REDDIT-BINARY). If CAMP is applied to large-scale graphs like OGB \cite{ogb}, then tangible pre-processing time will be consumed. We acknowledge that the CAMP framework is not scalable to large-scale graphs, which can be posed as a limitation of our approach.  

\section{Conclusion \& Future Works}
We designed an asynchronous message-passing framework, CAMP, that creates layer-specific node batches based on node centrality. Our proposed framework can be flexibly paired with a multitude of MP-GNNs. Furthermore, we also offer theoretical guarantees of the superiority of CAMP in mitigating oversquashing. We applied CAMP to six standard graph datasets to validate the effectiveness of the framework. The experimental results indicate that CAMP demonstrated impressive performance, underscoring the prowess of asynchronous message passing over structural rewiring. Exploring and designing asynchronous message-passing algorithms for dynamic graphs to alleviate oversquashing can be a potential future research avenue.

\bibliographystyle{IEEEtran}
\bibliography{ref}

\clearpage
\appendix

\section{Appendix}

\section*{Proofs}
\label{camp_theo_analysis}




\noindent
\textbf{Lemma \ref{lemma:camp-sensitivity}}
[Asynchronous Sensitivity Bound for CAMP]
Consider an $L$-layered GNN applied to a graph $G = (V, E)$ integrated with the CAMP framework under activation schedule
$\{C_n^{(\ell)}\}_{\ell=1}^{L}$. Assume that for every layer $\ell$, the Jacobians of $\Phi_\ell$ and $\Psi_\ell$ satisfy the Lipschitz bounds
\begin{equation}
    \left\|\nabla\Phi_\ell\right\| \leq \alpha, \qquad
    \left\|\nabla\Psi_\ell\right\| \leq \beta,
    \label{eq:lip}
\end{equation}
for real constants $\alpha, \beta > 0$.

\noindent\emph{(Part I: Sensitivity Bound.)}
For any pair of nodes $s, v \in V$,
\begin{equation}
    \left\|\frac{\partial x_v^{(L)}}{\partial x_s}\right\|_1
    \;\leq\;
    (\alpha\beta)^{K(s,v)}
    \sum_{\pi \in \mathcal{P}_{\mathrm{async}}(s,v)}
    \prod_{t=1}^{K(\pi)} S^{(\tau_t)}_{i_t,\, i_{t-1}}.
    \label{eq:camp_bound}
\end{equation}

\noindent\emph{(Part~II: Depth-wise Dominance over Synchronous
MP.)}
The synchronous sensitivity bound satisfies
\begin{equation}
    \left\|\frac{\partial x_v^{(L)}}{\partial x_s}
    \right\|_1^{\mathrm{sync}}
    \;\leq\;
    (\alpha\beta)^{L}\,(S^L)_{vs}.
    \label{eq:sync_bound}
\end{equation}
If $0 < \alpha\beta \leq 1$ and there exists at least one valid
asynchronous walk $\pi^* \in \mathcal{P}_{\mathrm{async}}(s,v)$ with
$K(\pi^*) = K(s,v) < L$, then
\begin{equation}
    \frac{\text{CAMP bound in \eqref{eq:camp_bound}}}
         {\text{Sync bound in \eqref{eq:sync_bound}}}
    \;\geq\;
    (\alpha\beta)^{K(s,v) - L}
    \;\xrightarrow{L\to\infty}\; +\infty,
    \label{eq:dominance}
\end{equation}
with $K(s,v)$ held fixed. In particular, for fixed graph distance
$d_G(s,v) = D$ and increasing network depth $L$, the synchronous
bound vanishes exponentially in $L$ while the CAMP bound's prefactor
$(\alpha\beta)^{K(s,v)}$ remains bounded below by $(\alpha\beta)^{D}$,
independent of $L$.
\begin{proof}
First, fix a source node $s \in V$. Define the scalar sensitivity
\[
    \delta_v^{(\ell)}(s)
    \;:=\;
    \left\|\frac{\partial x_v^{(\ell)}}{\partial x_s}\right\|_1.
\]

\noindent\textit{Case~1: $v \in C_n^{(\ell)}$ (node $v$ active at layer
$\ell$).}
Differentiating the active branch of \eqref{eq:camp_update} with respect to
$x_s$ via the chain rule:
\begin{align}
    \frac{\partial x_v^{(\ell+1)}}{\partial x_s}
    &= \nabla_1\Phi_\ell \cdot
       \frac{\partial x_v^{(\ell_v,\ell)}}{\partial x_s}
    \;+\; \nabla_2\Phi_\ell \sum_{u \in N(v)} S_{vu}^{(\ell)}
        \left(
            \nabla_1\Psi_\ell \cdot
            \frac{\partial x_v^{(\ell_v,\ell)}}{\partial x_s}
            + \nabla_2\Psi_\ell \cdot
            \frac{\partial x_u^{(\ell_u,\ell)}}{\partial x_s}
        \right).
    \label{eq:chain}
\end{align}
Taking the $\ell_1$ norm, applying the triangle inequality, and using the Lipschitz
bounds \eqref{eq:lip}, and $\sum_{u \in N(v)} S_{vu}^{(\ell)} \leq 1$
(row-normalization of $S^{(\ell)}$):
\begin{equation}
    \delta_v^{(\ell+1)}(s)
    \;\leq\;
    \alpha(1 + \beta)\,\delta_v^{(\ell_v,\ell)}(s)
    \;+\; \alpha\beta \sum_{u \in N(v)}
    S_{vu}^{(\ell)}\,\delta_u^{(\ell_u,\ell)}(s).
    \label{eq:rec_active}
\end{equation}

\noindent\textit{Case~2: $v \notin C_n^{(\ell)}$ (node $v$ inactive at
layer $\ell$).}
The inactive branch gives $x_v^{(\ell+1)} = x_v^{(\ell_v,\ell)}$,
so:
\begin{equation}
    \delta_v^{(\ell+1)}(s) = \delta_v^{(\ell)}(s).
    \label{eq:rec_inactive}
\end{equation}
Inactive nodes do \emph{not} accumulate any additional Lipschitz factor of $\alpha$ or $\beta$ at this layer, which is the fundamental structural difference from synchronous MP.

Let's define the diagonal activation indicator
$M^{(\ell)} = \mathrm{diag}(m_v^{(\ell)})$, where $m_v^{(\ell)} = 1$
if $v \in C_n^{(\ell)}$ and $0$ otherwise. Let
$\boldsymbol{\delta}^{(\ell)}(s) \in \mathbb{R}^n$ have entries
$\delta_v^{(\ell)}(s)$. Combining \eqref{eq:rec_active} and
\eqref{eq:rec_inactive}:
\begin{equation}
    \boldsymbol{\delta}^{(\ell+1)}(s)
    \;\leq\;
    \underbrace{%
        \Bigl[
            I - M^{(\ell)}
            + M^{(\ell)}\bigl(\alpha(1+\beta)I + \alpha\beta\,S^{(\ell)}\bigr)
        \Bigr]
    }_{=:\;T^{(\ell)}}
    \boldsymbol{\delta}^{(\ell)}(s),
    \label{eq:matrix_rec}
\end{equation}
where the self-term $\alpha(1+\beta)\delta_v^{(\ell_v,\ell)}$ is bounded
by $\alpha\beta$ (taking $\beta \geq 1$ without loss of generality for
nontrivial MESSAGE functions; for $\beta < 1$ replace $\alpha\beta$ by
$\alpha$ in the self-term, which only strengthens the bound). The
transition matrix $T^{(\ell)}$ has entries:
\begin{equation}
    T^{(\ell)}_{ij}
    = \begin{cases}
        \alpha\beta\,S^{(\ell)}_{ij}
          & i \in C_n^{(\ell)},\; j \neq i, \\
          \alpha(1 + \beta)
          & i \in C_n^{(\ell)},\; j = i, \\
        \mathbf{1}[i = j]
          & i \notin C_n^{(\ell)}.
    \end{cases}
    \label{eq:Tentry}
\end{equation}
Iterating \eqref{eq:matrix_rec} from layer $0$ with initial condition
$\boldsymbol{\delta}^{(0)}(s) = e_s$ (the standard basis vector, since
$\partial x_v^{(0)}/\partial x_s = \mathbf{1}[v=s]$):
\begin{equation}
    \boldsymbol{\delta}^{(L)}(s)
    \;\leq\;
    \left(\prod_{\ell=1}^{L} T^{(\ell)}\right) e_s.
    \label{eq:iterate}
\end{equation}

By the definition of matrix multiplication, we can expand the sum as follows:
\[
    \left(\prod_{\ell=1}^{L} T^{(\ell)}\right)_{vs}
    = \sum_{i_1, i_1, \ldots, i_{L-1}}
      T^{(1)}_{v,\, i_{L-1}}\,
      T^{(2)}_{i_{L-1},\, i_{L-2}}
      \cdots
      T^{(L)}_{i_1,\, s}.
\]
By \eqref{eq:Tentry}, each factor $T^{(\ell)}_{ab}$ is nonzero only when
(i) $a \in C_n^{(\ell)}$ and $(a,b) \in E$, contributing
$\alpha\beta\,S^{(\ell)}_{ab}$ (a genuine message step), or (ii) $a = b$,
contributing $1$ (a stay-step). A nonzero term therefore traces a
space-time path from $(s,0)$ to $(v,L)$ in $V \times [L]$ that either
stays at the same node (factor $1$) or moves along an active edge (factor
$\alpha\beta\,S^{(\ell)}$). Since $\{C_n^{(\ell)}\}$ partitions $V$,
each node is active at exactly one layer, so each node can be the
destination of a genuine traversal at most once. Stripping out stay-steps
and collecting genuine traversals, every nonzero path corresponds to a
valid asynchronous walk
$\pi = (i_0 = s, i_1, \ldots, i_K = v)
 \in \mathcal{P}_{\mathrm{async}}(s,v)$ with
$\tau_1 < \cdots < \tau_K$, giving:
\begin{equation}
    \left(\prod_{\ell=1}^{L} T^{(\ell)}\right)_{vs}
    = \sum_{\pi \in \mathcal{P}_{\mathrm{async}}(s,v)}
      (\alpha\beta)^{K(\pi)}
      \prod_{t=1}^{K(\pi)} S^{(\tau_t)}_{i_t,\, i_{t-1}}.
    \label{eq:expand}
\end{equation}
Substituting \eqref{eq:expand} into \eqref{eq:iterate} and reading off
the $v$-th entry establishes Part~I, equation~\eqref{eq:camp_bound}. In the following part, we will compare with the synchronous MP framework. 

\noindent\textit{Synchronous bound.}
In synchronous MP, $M^{(\ell)} = I$ for all $\ell$, so
$T^{(\ell)}_{\mathrm{sync}} = \alpha\beta(I + S)$. The same expansion
gives:
\begin{equation}
    \left(\prod_{\ell=1}^L
    T^{(\ell)}_{\mathrm{sync}}\right)_{vs}
    = (\alpha\beta)^L\,(S^L)_{vs},
    \label{eq:sync_expand}
\end{equation}
recovering the classical bound \eqref{eq:sync_bound}.

\medskip
\noindent\textit{Effect~1 (favorable for CAMP): reduction in contraction
depth.}
For every walk $\pi \in \mathcal{P}_{\mathrm{async}}(s,v)$,
$K(\pi) \leq L$. Taking $\pi^*$ of minimum length
$K(s,v) \leq d_G(s,v) =: D$, and using $0 < \alpha\beta \leq 1$ and
$K(s,v) \leq D \leq L$:
\begin{equation}
    (\alpha\beta)^{K(s,v)} \;\geq\; (\alpha\beta)^D \;\geq\;
    (\alpha\beta)^L.
    \label{eq:exp_compare}
\end{equation}
The prefactor in the CAMP bound is therefore strictly larger than in the
synchronous bound whenever $K(s,v) < L$.

\medskip
\noindent\textit{Effect~2 (unfavorable for CAMP): restriction of walks.}
Since
$\mathcal{P}_{\mathrm{async}}(s,v) \subseteq
 \mathcal{P}_{\mathrm{sync}}(s,v)$
and $S^{(\ell)}_{ij} \leq S_{ij}$:
\begin{equation}
    \sum_{\pi \in \mathcal{P}_{\mathrm{async}}(s,v)}
    \prod_{t=1}^{K(\pi)} S^{(\tau_t)}_{i_t, i_{t-1}}
    \;\leq\;
    \sum_{\pi \in \mathcal{P}_{\mathrm{sync}}(s,v)}
    \prod_{t=1}^{L} S_{i_t, i_{t-1}}
    = (S^L)_{vs}.
    \label{eq:path_compare}
\end{equation}

\medskip
\noindent\textit{Net effect.}
Effects~1 and~2 act in opposite directions. Define the depth-normalized
ratio
\[
    R(L) := \frac{(\alpha\beta)^{K(s,v)}}{(\alpha\beta)^{L}}
           = (\alpha\beta)^{K(s,v) - L}.
\]
Since $K(s,v) \leq D < L$ for sufficiently deep networks,
$K(s,v) - L < 0$ and $0 < \alpha\beta \leq 1$, so $R(L) > 1$ and
\begin{equation}
    R(L) = (\alpha\beta)^{K(s,v) - L}
    \;\xrightarrow{L\to\infty}\; +\infty,
    \label{eq:R_diverge}
\end{equation}
establishing \eqref{eq:dominance}. The practical implication is precise: in the CAMP bound, the effective contraction depth is $K(s,v) \leq d_G(s,v)$, independent of $L$. In synchronous MP, the contraction depth is always $L$, so $(\alpha\beta)^L(S^L)_{vs}$ vanishes exponentially in $L$ even when $d_G(s,v)$ is small. CAMP breaks this coupling because each node is activated at most once across all $L$ layers.
\end{proof}


\begin{corollary}
\label{cor:camp_depth}
    Under the conditions of Lemma~\ref{lemma:camp-sensitivity}, fix
    $s, v \in V$ with $d_G(s,v) = D$. Then for all $L \geq D$:
    \[
        \left\|\frac{\partial x_v^{(L)}}{\partial x_s}
        \right\|_1^{\mathrm{CAMP}}
        \;\leq\;
        (\alpha\beta)^{D}\,C_G(s,v),
    \]
    where
    \[
        C_G(s,v)
        = \sum_{\substack{\pi \in
          \mathcal{P}_{\mathrm{async}}(s,v) \\ K(\pi)=D}}
          \prod_{t=1}^{D} S^{(\tau_t)}_{i_t,i_{t-1}}
    \]
    is a constant depending only on the graph topology and the centrality
    partition, \emph{not} on $L$. In contrast, the synchronous bound
    satisfies
    \[
        \left\|\frac{\partial x_v^{(L)}}{\partial x_s}
        \right\|_1^{\mathrm{sync}}
        \;\leq\;
        (\alpha\beta)^{L}\,(S^L)_{vs} \;\to\; 0
        \quad \text{as } L \to \infty.
    \]
    Therefore, CAMP's sensitivity bound is bounded away from zero for
    all $L$, while the synchronous bound vanishes exponentially in $L$.
\end{corollary}

\begin{proof}
    In bound \eqref{eq:camp_bound}, retain only terms with
    $K(\pi) = D$ (the shortest valid asynchronous walks, which exist
    because the shortest path in $G$ between $s$ and $v$ becomes a valid
    asynchronous walk once its intermediate nodes are assigned in appropriate
    layers by the centrality partition). Dropping longer walks, which
    contribute non-negatively, gives the stated bound. The constant
    $C_G(s,v)$ is independent of $L$ by construction. 
    
    In the synchronous bound from \eqref{eq:sync_bound}, note that $(S^L)_{vs}$ does \emph{not} itself vanish as $L \to \infty$: for the symmetric normalized
    adjacency $S = D^{-1/2}AD^{-1/2}$, the vector $x$ with $x_i = \sqrt{d_i}$ satisfies $Sx = x$, so the spectral radius $\rho(S) = 1$ exactly for any connected graph (bipartite graph affects only the \emph{smallest} eigenvalue, not the largest). Instead, by standard finite Markov chain theory, $(S^L)_{vs} \to \pi_v$ as $L \to \infty$ for connected, non-bipartite $G$, where $\pi_v$ is the stationary distribution value
    at $v$ (proportional to $\deg(v)$) which is a nonzero constant. The term $(\alpha\beta)^L \to 0$ (given $0 < \alpha\beta < 1$) forces the product $(\alpha\beta)^L (S^L)_{vs}$ to zero, combined with $(S^L)_{vs}$ remaining bounded above by $1$ for all $L$ (a standard property of the normalized adjacency, since row sums are $1$). Hence
    $(\alpha\beta)^L (S^L)_{vs} \leq (\alpha\beta)^L \to 0$ as
    $L \to \infty$.
\end{proof}


\section*{Details of the Datasets}
\label{camp_data_details}

We considered six datasets collected from TUDatasets \cite{tudataset} and two Peptides graphs from LRGB \cite{lrgb}. The details of the datasets are provided in Table \ref{tab:tudatasets_details}. 

\begin{table*}[!ht]
    \centering
    \scriptsize
    \caption{Details of six datasets are provided, which are obtained from TUDatasets.}
    \label{tab:tudatasets_details}
    \begin{tabular}{l|ccccccc}
    \toprule
        Dataset & $\#$train & $\#$valid & $\#$test & $\#$avg nodes & $\#$avg edges & metric & node feats \\
        \midrule
        ENZYMES & $480$ & $60$ & $60$ & $32.63$ & $124.27$ & \text{accuracy} & \text{yes} \\
        MUTAG & $150$ & $18$ & $20$ & $17.93$ & $39.58$ & \text{accuracy} & \text{yes} \\
        PROTEINS & $890$ & $111$ & $112$ & $39.05$ & $145.63$ & \text{accuracy} & \text{yes}   \\
        COLLAB & $4000$ & $500$ & $500$ & $74.49$ & $4914.43$ & \text{accuracy} & \text{no} \\
        IMDB-BINARY & $800$ & $100$ & $100$ & $19.77$ & $193.66$ & \text{accuracy} & \text{no} \\
        REDDIT-BINARY & $1600$ & $200$ & $200$ & $429.62$ & $995.51$ & \text{accuracy} & \text{no} \\
        \bottomrule
    \end{tabular}
\end{table*}

\section*{Experimental Setup}
\label{camp_Exp_setup}

We executed experiments on the base GNN models coupled with CAMP on the $25$ random splits. For each split, the trained model is applied to the test set. The final test accuracy is presented with mean and standard deviation over all splits. The standard deviations are further scaled by the formula $s.d. \times \frac{2}{\sqrt{T}}$, where $T$ is the total number of random trials (here $T=25$), which is also followed in \cite{fosr}. Among the datasets, COLLAB, IMDB-BINARY, and REDDIT-BINARY do not have node features. As per the standard policy, we assigned ones as the features to every node.


\begin{table*}[!ht]
    \centering
    \scriptsize
    \caption{The hyperparameters for each dataset are provided to reproduce the best results. $L$ denotes the number of message-passing layers of the underlying GNN.}
    \label{tab:camp_hyperparams}
    \begin{tabular}{lcccccccccccc}
    \toprule
    \multirow{2}{*}{Model} & \multicolumn{2}{c}{ENZYMES} & \multicolumn{2}{c}{MUTAG} & \multicolumn{2}{c}{PROTEINS} & \multicolumn{2}{c}{COLLAB} & \multicolumn{2}{c}{IMDB-BINARY} & \multicolumn{2}{c}{REDDIT-BINARY} \\
    \cmidrule{2-13}
    & L & Centrality & L & Centrality & L & Centrality & L & Centrality & L & Centrality & L & Centrality \\
    \midrule
    GCN & 10 & Closeness & 16 & Degree & 16 & Degree & 10 & Load & 12 & Pagerank & 16 & Closeness \\
    GIN & 10 & Closeness & 12 & Degree & 16 & Closeness & 12 & Betweenness & 10 & Pagerank & 12 & Closeness \\
    \bottomrule
    \end{tabular}
\end{table*}

\section*{Hyperparameter Details}
\label{camp_hyper_details}

The models are trained with a learning rate of $0.001$ and a dropout rate of $0.50$. Model parameters are optimized with Adam, and weight decay is fixed at $10^{-5}$. In each experiment, we employed both $4$-layer GCN and GIN models with a batch size of $64$. These are the fixed hyperparameters for the entire experiment. The other data-specific hyperparameters are provided in Table \ref{tab:camp_hyperparams}.

\section*{Dirichlet Energy}
\label{camp_de_details}

Let us consider graph $G$ with adjacency matrix $A$ and symmetrically normalized graph Laplacian $\Tilde{L} = \text{I} - \Tilde{D}^{-\frac{1}{2}} \Tilde{A} \Tilde{D}^{-\frac{1}{2}}$, then the Dirichlet energy \cite{chung_spectral} of the set of features $X$ with Laplacian is represented as following, 
\begin{equation}
\label{camp_de_supple_eq}
\begin{split}
    \text{DE}(X, \Tilde{L}) &= \text{Tr} (X^{\top} \Tilde{L} X) \\
    &= \frac{1}{2} \sum\limits_{(i, j) \in \mathcal{E}} A_{ij} \left( \frac{X_i}{\sqrt{d_i}} - \frac{X_j}{\sqrt{d_j}} \right)^{2}
\end{split}
\end{equation}
The Dirichlet energy is directly connected with the measure of oversmoothing in GNNs. The average value of DE indicates that the node features became indistinguishable when neighborhood aggregation was performed in the deeper layers. In our experiments, we utilized Eq \ref{eq:camp_de_supple_eq} to compute the energies at the final layer of the architectures.

\section*{Signal Propagation and Total Effective Resistance}
\label{camp_signal_prop}

We will present the estimation of signal propagation across the graph with respect to the normalized total effective resistance as discussed in the Experiments. For any input graph, first, consider a source node $v$ and assign a $p$-dimensional feature vector with zero vectors assigned to the rest of the nodes. Initialize a model randomly with $m$ layers; then the amount of signal propagated can be estimated as,  
\begin{equation}
\label{camp_signal_supple_eq}
    h^{(m)}_{\odot} = \frac{1}{p \max_{u \neq v} d_{\mathcal{G}}(u, v)} \sum\limits_{j=1}^{p} \sum\limits_{u \neq v} \frac{h^{(m),j}}{\lVert h^{(m),j} \rVert } d_{\mathcal{G}}(u, v)
\end{equation}
where $h^{(m),j}$ denotes the $j^{th}$ component of the feature at the $m^{th}$ layer, $d_{\mathcal{G}}(u, v)$ is the shorted distance between the node $u$ and $v$. The normalized signal $\frac{h^{(m),j}}{\lVert h^{(m),j} \rVert }$ weighted by $d_{\mathcal{G}}(u, v) / \max_{u \neq v} d_{\mathcal{G}}(u, v)$ is propagated from source node $v$ to all other nodes. The final signal is averaged over the number of feature dimensions $p$. In our context, for each graph, we selected $10$ source nodes and the final signal $h^{(m)}_{\odot}$ with total effective resistance averaged over each source node. This process is continued over all graphs, and total effective resistance and signal propagation are plotted.

\section*{Additional Experimental Results}

\subsection*{Effect of Centrality on the Performance of CAMP-GIN}
\label{camp_centrality_gin_subsection}

We present a comparative study of the performance of CAMP-GIN in Table \ref{tab:centrality_gin}. The experiment was conducted on a $10$-layered GIN architecture. The optimal performances for each dataset are attained for different centrality measures, highlighting the importance of centrality choices. 

\begin{table*}[!ht]
\centering
\scriptsize
\caption{A comparative study of the performance of CAMP-GIN is presented based on five different centrality measures.}
\label{tab:centrality_gin}
\begin{tabular}{lcccccc}
\toprule
Centrality & ENZYMES & MUTAG & PROTEINS & COLLAB & IMDB-BINARY & REDDIT-BINARY \\
\midrule
Degree & $43.13\pm2.88$ & $\mathbf{80.00\pm4.36}$ & $72.39\pm1.89$ & $72.34\pm0.77$ & $70.64\pm1.68$ & $75.38\pm1.13$ \\
Betweenness & $41.87\pm2.62$ & $77.00\pm4.16$ & $72.07\pm2.02$ & $\mathbf{73.30\pm0.86}$ & $69.76\pm2.15$ & $72.70\pm0.90$ \\
PageRank & $41.00\pm2.54$ & $74.00\pm4.08$ & $72.68\pm1.55$ & $72.66\pm0.79$ & $\mathbf{71.48\pm1.47}$ & $77.98\pm1.37$ \\
Load & $39.93\pm2.44$ & $76.00\pm4.28$ & $72.64\pm2.26$ & $72.97\pm0.85$ & $69.84\pm1.94$ & $72.82\pm0.78$ \\
Closeness & $\mathbf{46.60\pm2.26}$ & $76.20\pm2.96$ & $\mathbf{74.07\pm1.73}$ & $72.57\pm0.61$ & $71.00\pm1.99$ & $\mathbf{81.98\pm1.20}$ \\
\bottomrule
\end{tabular}
\end{table*}

\subsection*{Effect on Dirichlet Energy}
\label{camp_de_subection}

\begin{figure*}[!ht]
    \centering
    \includegraphics[width=\textwidth]{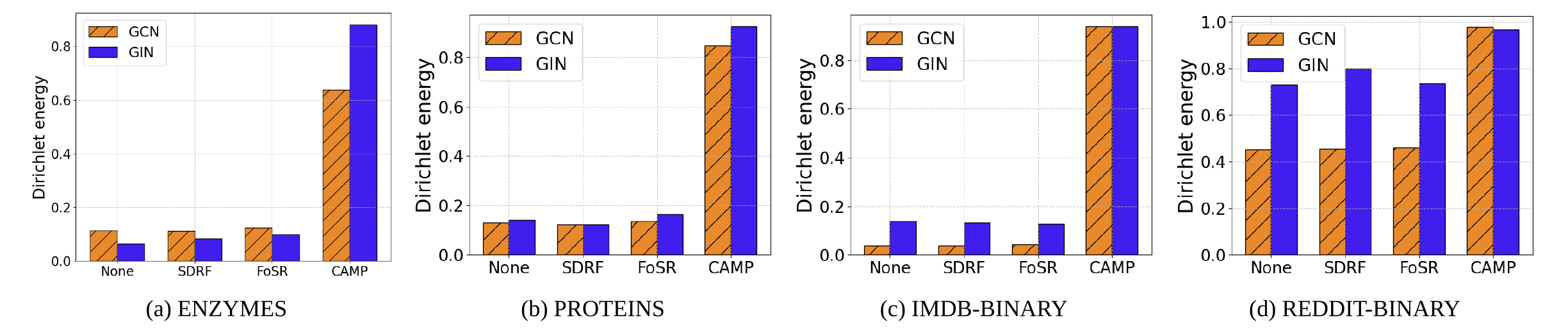}
    \caption{The Dirichlet energies of the last layer of the models are presented. CAMP attains better energy values than other rewiring methods. }
    \label{fig:camp_eenrgy_supple}
\end{figure*}

The additional results are presented in Figure \ref{fig:camp_eenrgy_supple}, where CAMP produces higher Dirichlet energy compared to other rewiring methods. Thus, CAMP showed the capacity to prevent feature collapse in multi-layered GNNs. 

\subsection*{Effect on Oversmoothing}
\label{camp_oversmooth_subsection}

\begin{figure*}[!ht]
    \centering
    \includegraphics[width=\textwidth]{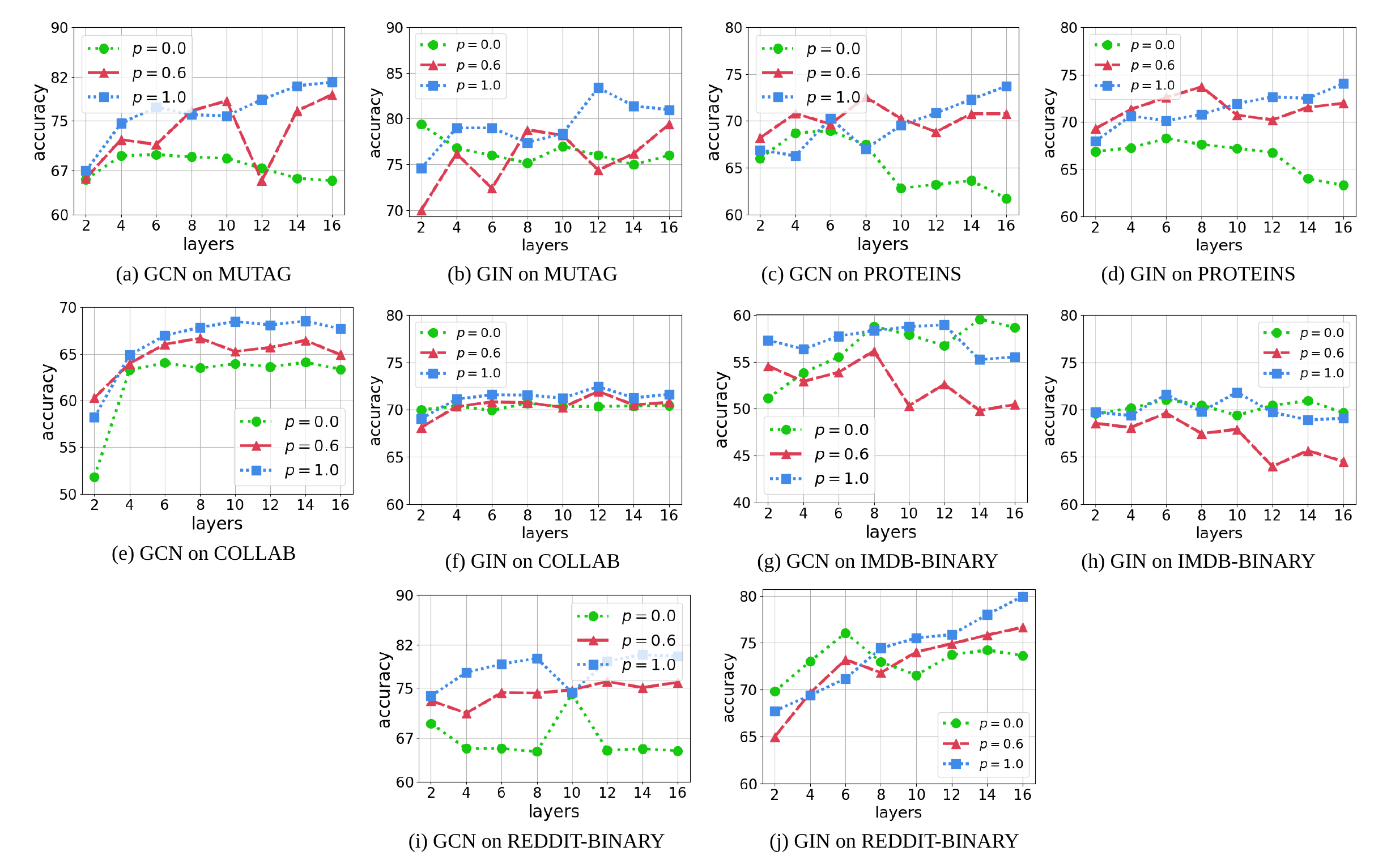}
    \caption{Performance of CAMP with different batch sampling rates is presented for various network depths. The test accuracy is observed when the full subset is selected as the node batches. }
    \label{fig:camp_ovrsmooth_supple}
\end{figure*}

The additional plots for the performance of CAMP in deeper layers are presented in Figure \ref{fig:camp_ovrsmooth_supple}. The performance trends suggest the power of CAMP in tackling oversmoothing in deep GNNs. 

\subsection*{Signal Propagation}
\label{camp_signal_prop}

\begin{figure*}[!ht]
    \centering
    \includegraphics[width=\textwidth]{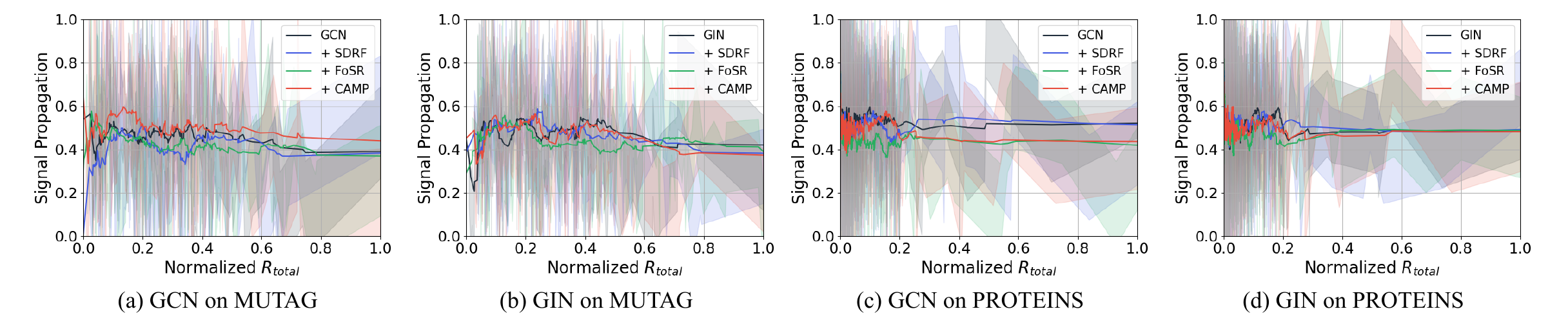}
    \caption{The signal propagation with respect to normalized $R_{\text{out}}$ is presented for MUTAG and PROTEINS. CAMP exhibited competitive performance compared to other rewiring methods by propagating a steady signal across the network with increasing resistance.  }
    \label{fig:signal_prop_supple}
\end{figure*}

The additional results pertaining to the efficiency of CAMP in propagating the signal are presented in Figure \ref{fig:signal_prop_supple}. Our proposed framework, CAMP, demonstrated steady signal strength with increasing effective resistance of the graphs. 

\subsection*{Effect of Batch Variation}
\label{camp_batch_subsection}

\begin{figure*}[!ht]
    \centering
    \includegraphics[width=\textwidth]{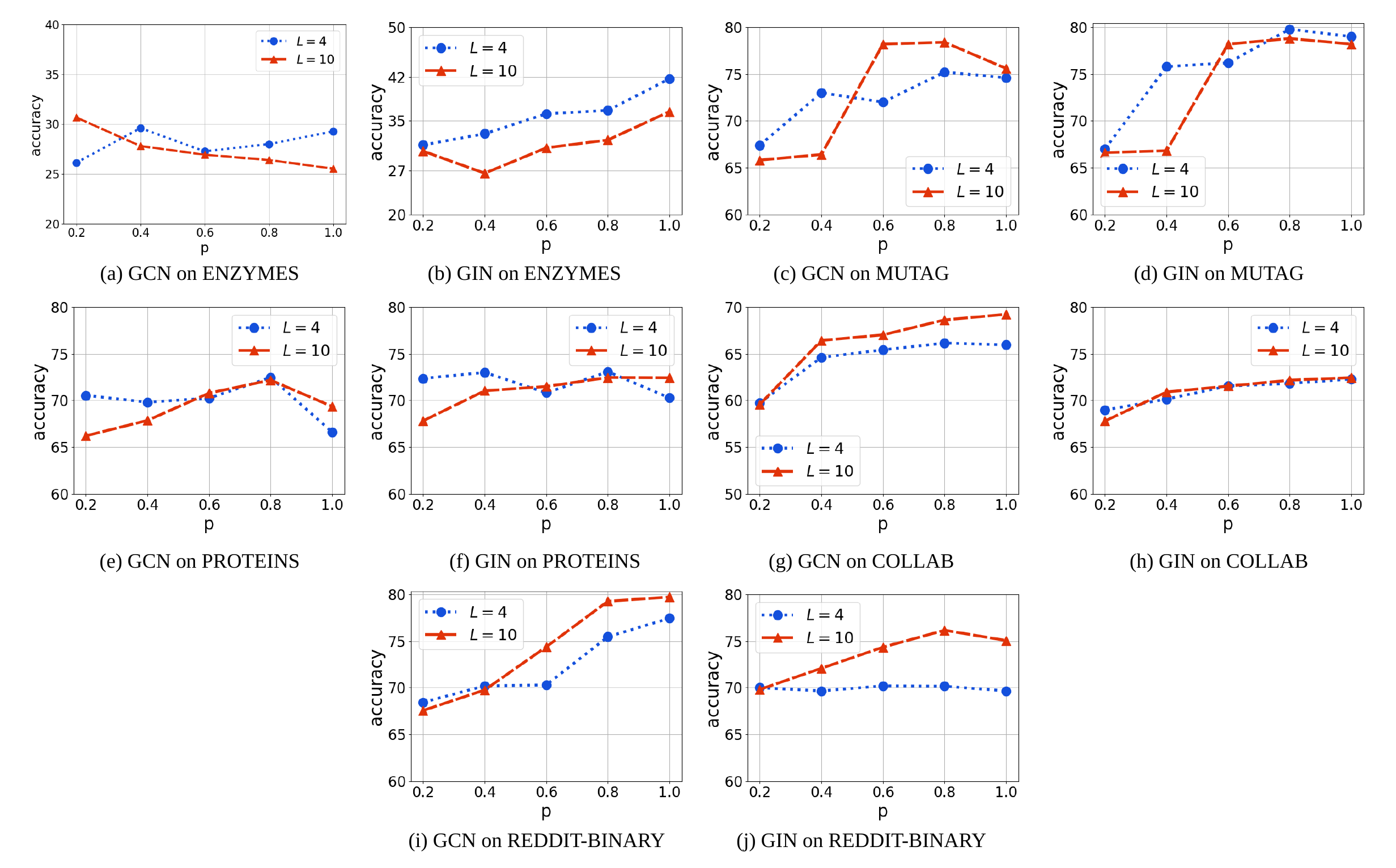}
    \caption{The performance variation is illustrated with various node batch sampling rates. The best results are mostly obtained when the complete subset is chosen as a node batch for the hidden layers. }
    \label{fig:camp_batch_supple}
\end{figure*}

The additional results of the performance of GCN and GIN paired with CAMP are presented over various sampling probabilities. The experiments are performed for two different layers $4$ and $10$. The plots are shown in Figure \ref{fig:camp_batch_supple}. CAMP achieved the best performance in most cases when the sampling probability is $1$, implying that selecting the full subset in a particular layer improves message passing effectively. 

\end{document}